\documentclass[sigconf,10pt]{acmart}

\usepackage{amsmath}
\usepackage{algorithm}
\usepackage{algorithmic}
\usepackage{graphicx}
\usepackage{lipsum}
\usepackage{enumerate}
\usepackage{enumitem}
\usepackage{hyperref}
\usepackage{multirow, makecell}
\usepackage{nccmath}
\usepackage{ragged2e}
\usepackage[caption=false,font=normalsize]{subfig}
\usepackage{threeparttable} 
\usepackage{url}
\usepackage{xspace}
\usepackage{CJKutf8}
\usepackage{bbding}
\usepackage{color}
\usepackage{tabularray}
\usepackage{graphicx}
\usepackage{booktabs}
\usepackage{rotating}
\usepackage{multirow}

\AtBeginDocument{%
  \providecommand\BibTeX{{%
    \normalfont B\kern-0.5em{\scshape i\kern-0.25em b}\kern-0.8em\TeX}}}

\setcopyright{acmcopyright}
\copyrightyear{2024}
\acmYear{2024}
\acmDOI{00000.00000}

\acmConference[ACM MobiSys ’24]{In The 22nd ACM International Conference on Mobile Systems, Applications, and Services}{June 03--07,
  2024}{Tokyo, Japan}

\newcommand{\ie}{\emph{i.e.},\xspace}

\newcommand{\eg}{\emph{e.g.},\xspace}
\newcommand{\etc}{\emph{etc.}\xspace}

\newcommand\figref[1]{Fig.~\ref{#1}}

\newcommand\tabref[1]{Tab.~\ref{#1}}

\newcommand{\fakeparagraph}[1]{\vspace{1mm}\noindent\textbf{#1.}}

\newcommand{\shortauthor}{Liu et al.}

\ifodd 1

\else

\fi

\begin{document}
\begin{CJK}{UTF8}{gbsn}

%
\title{Deep Learning Inference on Heterogeneous Mobile Processors: Potentials and Pitfalls}
\author{Sicong Liu}
\orcid{0000-0003-4402-1260}
\affiliation{%
  \institution{Northwestern Polytechnical University}
  \country{}
}

\author{Wentao Zhou}
\orcid{0009-0002-3165-4288}
\affiliation{%
  \institution{Harbin Engineering University}
  \country{}
}

\author{Zimu Zhou}
\orcid{0000-0002-5457-6967}
\affiliation{%
  \institution{City University of Hong Kong}
  \country{}
}

\author{Bin Guo$^{\ast}$}
\orcid{0000-0001-6097-2467}
\affiliation{%
  \institution{Northwestern Polytechnical University}
  \country{}
  \thanks{*Corresponding author: guob@nwpu.edu.cn}
}

\author{Minfan Wang}
\orcid{0009-0002-0046-4828}
\affiliation{%
  \institution{Chang'an University}
  \country{}
}

\author{Cheng Fang}
\orcid{0009-0000-8131-6178}
\affiliation{%
  \institution{Northwestern Polytechnical University}
  \country{}
}

\author{Zheng Lin}
\orcid{0009-0006-4977-1385}
\affiliation{%
  \institution{Northwestern Polytechnical University}
  \country{}
}

\author{Zhiwen Yu}
\orcid{0000-0002-9905-3238}
\affiliation{%
  \institution{Harbin Engineering University}
  \country{}
}
\affiliation{%
  \institution{Northwestern Polytechnical University}   
  \country{}
}

\begin{abstract}
There is a growing demand to deploy computation-intensive deep learning (DL) models on resource-constrained mobile devices for real-time intelligent applications. 
Equipped with a variety of processing units such as CPUs, GPUs, and NPUs, the mobile devices hold potential to accelerate DL inference via parallel execution across heterogeneous processors.
Various efficient parallel methods have been explored to optimize computation distribution, achieve load balance, and minimize communication cost across processors.
Yet their practical effectiveness in the dynamic and diverse real-world mobile environment is less explored. 
This paper presents a holistic empirical study to assess the capabilities and challenges associated with parallel DL inference on heterogeneous mobile processors. 
Through carefully designed experiments covering various DL models, mobile software/hardware environments, workload patterns, and  resource availability, we identify limitations of existing techniques and highlight opportunities for cross-level optimization.

\end{abstract}



\maketitle

\section{Introduction}
\label{sec:intro}

The deployment of deep learning (DL) models has shifted from cloud-centric to mobile devices for on-device intelligence~\cite{ song2022integrating,liu2022real, guan2022deepmix, arrotta2022dexar}. 
This transition enables various applications that interact intelligently with users in \textit{real-time}, including biometric authentication on smartphones~\cite{song2022integrating}, arm posture tracking on smartwatches~\cite{liu2022real}, 3D object detection on headsets~\cite{guan2022deepmix}, and language translation on home devices~\cite{arrotta2022dexar}.

Commercial mobile system-on-chips (SoCs) increasingly integrate a variety of processing units, such as CPUs, GPUs, DSPs, and NPUs, enabling an emerging trend towards accelerating DL inference through \textit{parallel} execution across \textit{heterogeneous} resources~\cite{kim2019mulayer,jia2022codl,wei2023nn,jeong2022band}. 
It involves mapping DL computations to the processors best suited for their execution, which takes place at two levels (see \figref{fig:overview}). 
\textit{i)}
At the \textit{resource-friendly frontend compilation level}, a DL model is compiled into a computational directed acyclic graph (DAG). 
\textit{ii)}
At the \textit{model-adaptive backend compilation level}, the DAG operators are then mapped and scheduled across heterogeneous computing/memory resources for parallel execution.
As mainstream mobile DL frameworks typically restrict computations to a \textit{single} processor at a time~\cite{tensorflowlite,mace,ncnn}, existing research developed various strategies to achieve \textit{cross-processor} parallel inference speedup across heterogeneous processors, including \textit{i)} optimizing the distribution of computations by branching and partitioning subgraphs \cite{wei2023nn}, layers \cite{kim2019mulayer}, or operators \cite{jia2022codl} among processors; \textit{ii)} balancing loads across processors to minimize asynchronous waiting times \cite{jeong2022band}; and \textit{iii)} reducing data communication overhead associated with merging results from concurrent executions \cite{zhang2023edgenn}.

Despite prior efforts on the feasibility of parallel inference across processors, their efficacy in \textit{real-world mobile contexts} remains less-explored. 
We hypothesize that backend compilation-level optimization for maximal utilization of heterogeneous resources, as many existing techniques do~\cite{kim2019mulayer,jia2022codl,wei2023nn}, may not yield optimal overall performance due to the \textit{diversity} and \textit{dynamics} of the mobile device ecosystem.
By evaluating representative techniques across different DL models, mobile software/hardware environments, workload patterns, and runtime resource availability, our empirical study strives to answer the following questions when applying these techniques in practice.

\textit{(I) Are existing strategies for parallel inference across mobile processors adequately effective?}
    For previous solutions focused on \textit{backend compilation-level} optimization,
    our study implies that their \textit{design space} could be improved both \textit{within} and \textit{beyond} the current scope.
    \vspace{-3mm}
    \begin{itemize}
        \item \textit{Unsupported Operators and Processor Fallbacks}: 
        The operators generated at the backend compilation-level optimization are not supported by all processors, leading to operational fallbacks. 
        For instance, when GPUs encounter unsupported operators, the DL computations are forced to revert to CPUs, leading to suspensions of processes and under-utilization of resources. 
            Our experiments show that unsupported operators \textit{vary} across heterogeneous mobile processor combinations, with significant fallbacks in certain cases \eg for NN-Stretch in ResNet-50 on the Xiaomi 9, the ratio of unsupported operators during parallel inference on the CPU and GPU reaches approximately 48\%.
        \item \textit{Cross-Level Optimization Opportunities}:
        Existing solutions agnostic to DL models often exhibit \textit{overparameterization} that can be optimized before DAG conversion. 
        Our experiments demonstrate that incorporating frontend compilation-level optimizations, such as pruning, can improve resource utilization and accelerate DL inference at the backend.
        Also, optimizing operator fusion/parallelism and memory swap/allocation at the model-adaptive backend compilation level can enhance data reuse and eliminate memory segmentation, broadening the scope of frontend optimization.
        This indicates a potential to expand the design space via cross-level optimization, prioritizing the parallel execution of crucial operations while selectively managing less critical ones, optimizing the system scheduling for higher concurrency and resource availability.
    \end{itemize}
    \textit{(II) Is parallel inference across processors always beneficial?}
    Most strategies maximize the resource utilization across processors for DL task alone.
    Such \textit{optimization objectives} may not fit in real-world mobile contexts. 
    \begin{itemize}
        \item \textit{Granularity of Parallel Scheduling}:
        On mobile heterogeneous multiprocessor devices, a too \textit{large} scheduling granularity can lead to idle chip cores, undermining the efficiency of utilizing multiple processors.
        In contrast, an overly \textit{fine} granularity may cause process suspensions and blocks due to increased data transmission delays. 
        Furthermore, our experiment suggest that allocating computations to a single processor could, in some instances, outperform distribution across heterogeneous processors due to scheduling overhead.
        \item \textit{Presence of Competing Processes}: 
        Achieving maximum resource utilization for DL inference does not necessarily result in optimal overall system performance, particularly in multi-process mobile applications.
        Dedicating excessive computational resources to DL inference can impair the functionality of other components, \eg UI responsiveness in smartphones and AR apps.
        Given the \textit{dynamic nature} of runtime resource availability and competing process demands in mobile applications, pinpointing optimal resource utilization levels is essential for improving the overall performance without compromising other processes.
    \end{itemize}
Our main contributions are as follows.
\begin{itemize}
    \item 
    We conduct pilot studies tailored to evaluate the performance of parallel DL inference strategies across heterogeneous mobile processors in real-world conditions.
    They assess representative solutions in setups critical to practical applications, yet under-explored in existing empirical studies.
    \item 
    We identify the limitations and opportunities in the design space and objectives of existing solutions.
    These insights would facilitate effective parallel inference across heterogeneous resources amid the inherent diversity and dynamics of the mobile device ecosystem.
\end{itemize}

\section{Methodology}
\label{sec:Methodology}

\begin{figure}[h!]
  \centering
  \includegraphics[width=0.73\linewidth]{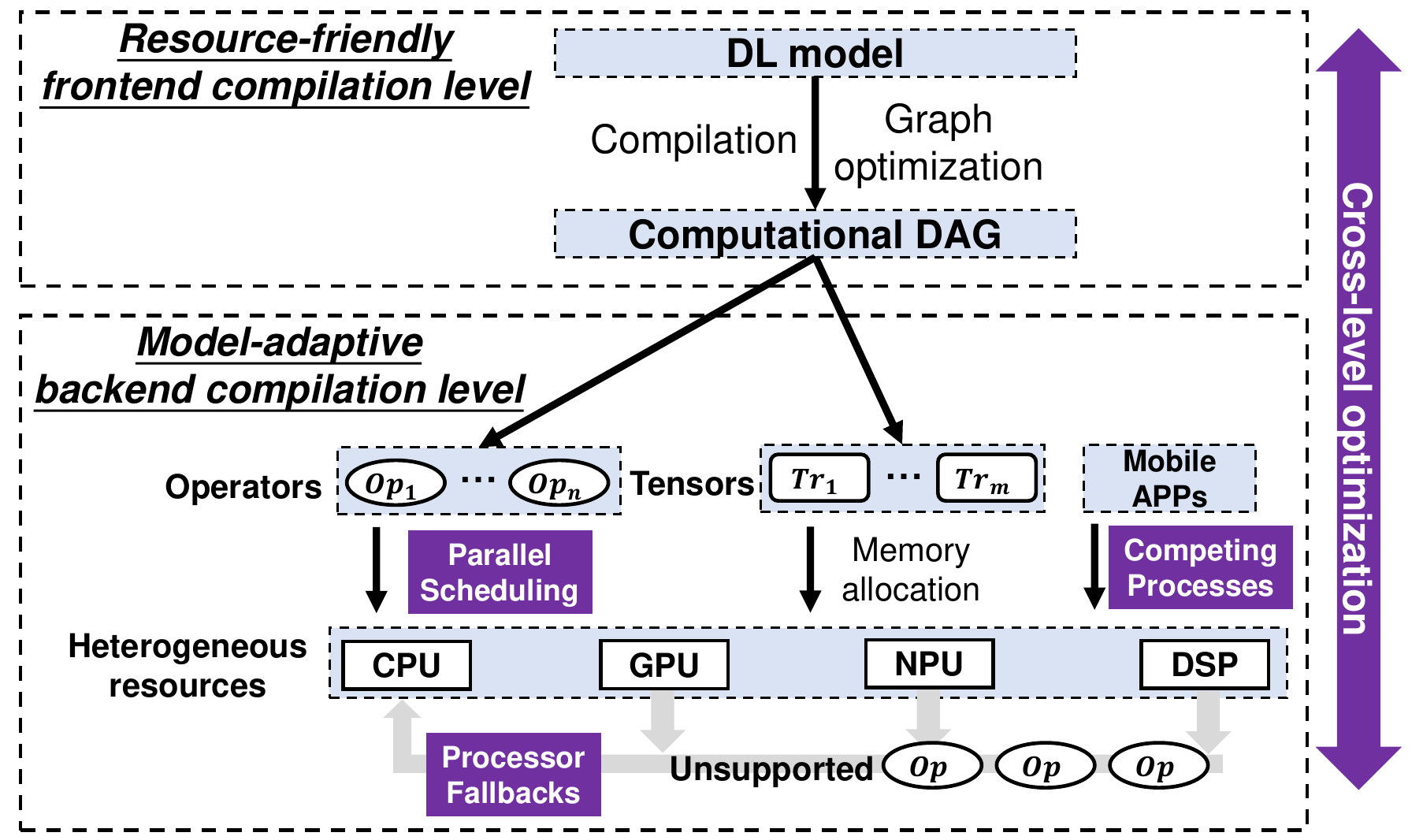}
  \vspace{-3mm}
  \caption{Illustration of parallel DL inference workflow across heterogeneous processors on mobile devices.}
  \vspace{-3mm}
  \label{fig:overview}
\end{figure}

\fakeparagraph{Optimization Stack} 
As previously mentioned, parallel DL inference across heterogeneous processors can be optimized at two levels, \ie frontend and backend compilation.
\figref{fig:overview} illustrates the generic workflow to map the DL computation to different resources. 
\textit{(i)} \textit{Frontend compilation} translates DL models into an intermediate representation, \ie \textit{computational graph}, eliminating redundant operators in the graph.
\textit{(ii)} \textit{Backend compilation} further compiles the \textit{computational graph} into an executable binary file, involving three sequential processes: 
\textit{\textcircled{1} Graph optimization} merges operators that conform to fusion rules and schedule the computation graph for parallel execution based on four levels: intra-operator, inter-operator, sub-graph, and task.
\textit{\textcircled{2} Operator distribution} assigns each operator to processors based on operator types and hardware support. 
\textit{\textcircled{3} Memory allocation} allocates memory space for data in the graph based on operator allocation and the available memory resource of processors.
The \textit{dynamics} in mobile resource availability, system constraints, workload, and input data can influence the performance of generic compilation optimization strategies.

\fakeparagraph{Existing Strategies}
To our best knowledge, no solution covers the entire optimization stack above.
We list representative resource-efficient DL inference methods in two categories based on our generic compilation workflow.

\begin{itemize}
    \item \textit{Frontend compilation} optimization level reduces the redundancy of DL models to minimize resource demand.
    \begin{itemize}
        \item \textbf{Pruning}~\cite{fan2023sparse} removes redundant DL model parameters, channels, and connections.
        \item \textbf{Low-rank decomposition}~\cite{yin2021towards} adopts SVD to use basis vectors to represent the convolution kernel.
        \item \textbf{Parameter/activation quantization}~\cite{wang2023spaceevo} represents 32-bit parameters with 8-bit widths.
    \end{itemize}
    \item \textit{Backend compilation} optimization level improves hardware resource availability and utilization.
    \begin{itemize}
        \item \textbf{Operator fusion} ~\cite{niu2021dnnfusion} fuses adjacent operators into a new operator to save  I/O of intermediate results.
        \item \textbf{Mace}~\cite{mace}: \textit{single-processor} inference strategy.
        \item \textbf{TensorFlow Lite}~\cite{tensorflowlite}: \textit{single-processor}  strategy.
        \item \textbf{NN-Stretch}~\cite{wei2023nn}: converts DL models into multi branches to increase computation parallelism.
        \item \textbf{$\mu$Layer}~\cite{kim2019mulayer}: divides DL layers into channels and executes diverse channels across processors.
        \item \textbf{CoDL}~\cite{jia2022codl}: partitions DL model operators into operator chains for operator parallel execution.
        \item \textbf{Memory allocation} ~\cite{symons2021loma} releases memory in time.   
    \end{itemize}
\end{itemize}

\fakeparagraph{Overall Experimental Setups}
We aim to assess the limitations and opportunities of the above methods for parallel DL inference across heterogeneous processors, with a special focus on the \textit{diversity} and \textit{dynamics} of mobile devices.
We experiment with 8 commercial mobile devices equipped with heterogeneous multiprocessors (see \tabref{tab:devices-table}), and simulate various \textit{internal} and \textit{external} system factors affecting the runtime resource availability on these mobile devices (\eg DL workloads, DVFS strategies, competing processes).
In addition to accuracy and latency, we also include multiple intermediate performance metrics such as the reduction in DL resource demands (\eg model size, memory usage), and the improvement in runtime resource utilization (\eg utilization rate, cache hit rate, and frame drop rate).
As with other empirical studies and benchmarks ~\cite{zhang2022comprehensive, wang2020benchmarking, feng2022benchmark, luo2020comparison}, we mainly test DL models for \textit{visual} related tasks, covering representative models including YOLOv2 \cite{redmon2017yolo9000}, VGG-16 \cite{simonyan2014very}, PoseNet \cite{zimmermann2017learning}, Fast Style Transfer \cite{an2019fast}, RetinaFace \cite{deng2020retinaface}, ResNet-18/50/34 \cite{he2016deep} and RegNetX-1.6GF/4GF \cite{radosavovic2020designing}.

\begin{table}[t]
\centering
\caption{Specifications of mobile devices for evaluation.}
\vspace{-3mm}
\label{tab:devices-table}
\resizebox{0.73\columnwidth}{!}{%
\begin{tabular}{|c|c|c|c|c|}
\hline
\textbf{Mobile devices} &
  \textbf{SoC} &
  \textbf{Processor conf.} &
  \textbf{RAM} \\ \hline
\textbf{$D_1$: Xiaomi 9} &
  Snapdragon 855 &
  CPU+GPU+DSP &
  8GB\\ \hline
\textbf{$D_2$: Huawei Nova 7} &
  Kirin 985 &
  CPU+GPU+DSP+NPU &
  8GB\\ \hline
\textbf{$D_3$: Redmi K30 Pro} &
  Snapdragon 865 &
  CPU+GPU+DSP &
  6GB\\ \hline
\textbf{$D_4$: Xiaomi 11 Pro} &
  Snapdragon 888 &
  CPU+GPU+DSP &
  8GB\\ \hline
\textbf{$D_5$: Honor V30 Pro} &
  Kirin 990 &
  CPU+GPU+DSP+NPU &
  8GB\\ \hline
  \textbf{$D_6$: Honor 9} &
  Kirin 960 &
  CPU+GPU+DSP &
  6GB\\ \hline 
\textbf{$D_7$:  Huawei Matepad Pro} & 
Snapdragon 870 & 
CPU+GPU+DSP &
6GB \\ \hline
\textbf{$D_8$: Pixel 6} & 
Google Tensor & 
CPU+GPU &
8GB \\ \hline
\end{tabular}%
\vspace{-6mm}
}
\vspace{-6mm}
\end{table}

\section{Experimental Results}
\label{sec:Performance Analysis}
\subsection{Are existing strategies for parallel inference adequately effective? }

\begin{figure*}[t]
  \centering
  \includegraphics[width=0.66\textwidth]{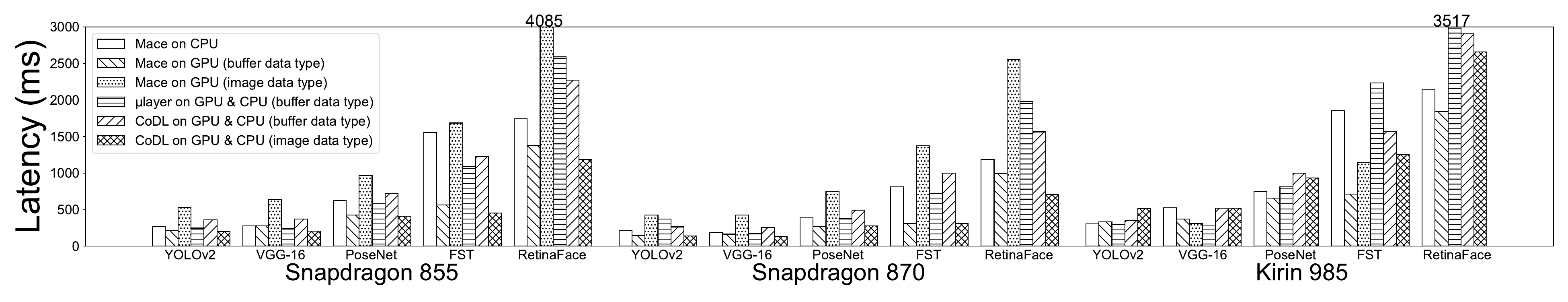}
  \vspace{-3mm}
  \caption{Latency on different mobile devices with diverse parallel inference strategies.}
  \vspace{-3mm}
  \label{fig:latency}
\end{figure*}

\subsubsection{Performance of different parallel strategies over diverse mobile devices}
We compare the inference latency of six strategies: \textcircled{1} Mace framework executes inference on CPU, \textcircled{2} Mace on GPU with the buffer type, \textcircled{3} Mace on GPU with the image type, \textcircled{4} $\mu$Layer on CPU+GPU, \textcircled{5} CoDL on CPU+GPU in parallel with buffer type, and \textcircled{6} CoDL on CPU $\&$ GPU in parallel with the image type.
We test them on five models, \ie YOLOv2, VGG-16, PoseNet, Fast Style Transfer (FST), and RetinaFace, across three mobile devices, \ie Snapdragon 855 ($D_1$), Snapdragon 870 ($D_7$), and Kirin 985 ($D_2$).
Figure \ref{fig:latency} shows the result.
\textit{First}, 
different parallel strategies exhibit varying speedup performance across diverse DL models and mobile devices. 
On $D_1$, CoDL achieves optimal speedup for parallel inference using image data type, while on $D_2$, its latency compared to running on a single GPU ranges from $1.4\times$ to $1.8\times$. 
This stems from variations in buffer and image data type support in Adreno and Mali architecture GPUs.
\textit{Second}, executing inference on a single processor may outperform distribution across heterogeneous processors in certain cases. 
Parallel inference of FST on CPU+GPU using $\mu$Layer slows down by $1.9\times \sim 3.1\times$ compared to on GPU, due to scheduling overhead.

\textbf{Summary}. The performance of various parallel strategies on diverse models varies. 
Even within the same parallel strategy, performance diverges. 
There's \textit{no one-fit-all} parallel strategy,  
adaptation is essential to enhance resource utilization across diverse mobile platforms and models.

\begin{figure}[t]
  \centering
  \includegraphics[width=0.3\textwidth]{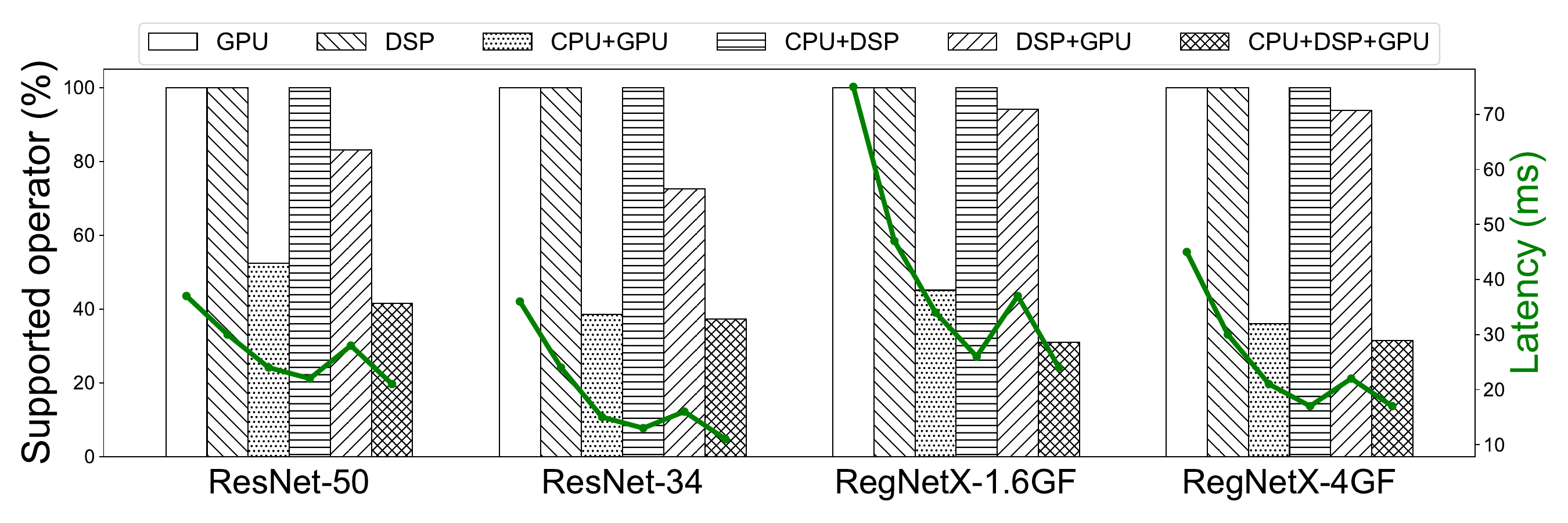}
  \vspace{-3mm}
  \caption{Operator support over diverse models.}
  \label{fig:operator_supported}
\end{figure}

\subsubsection{Challenges in unsupported operators and process fallbacks.} 
\label{subsubsec_fallback_phenomenon}
To test the presence of unsupported operators in existing parallel inference strategies, we use the Snapdragon 855 ($D_1$) to evaluate six inference schemes across four diverse DL models: \ie single-processor inference on GPU, DSP using TensorFlow Lite, as well as parallel inference on CPU+GPU, CPU+DSP, CPU+GPU, and CPU+GPU+DSP using NN-stretch.
Figure \ref{fig:operator_supported} shows the results.
\textit{First}, unsupported operators are ubiquitous across diverse parallel schemes and DL models.
The presence of unsupported operators triggers process fallback, causing computations to regress from specialized accelerators like GPU, DSP, \etc to the CPU.
\textit{Second}, the operator fallback leads to resource idleness and process waiting. 
In ResNet-50, the unsupported operator ratio of parallel inference on CPU and GPU reaches about 48\%. 
We note that cross-processor parallel inference introduces new operators, decreasing the percentage of supported operators.

\textbf{Summary}. 
At the backend compilation level, considering operator support for cross-processor distribution is crucial. Also, adaptive redistribution can enhance underutilized processor resources in cases of process fallback at runtime.

\begin{table}[t]
\caption{Performance of cross-level optimization of parallel inference, \ie CoDL, on Snapdragon 855.}
\vspace{-3mm}
\label{tab:cross}
\resizebox{\columnwidth}{!}{
\begin{tabular}{|c|c|c|c|c|c|}
\hline
\textbf{Levels} & \textbf{Methods} & \textbf{\begin{tabular}[c]{@{}c@{}}Top\\ accuracy (\%)\end{tabular}} & \textbf{\begin{tabular}[c]{@{}c@{}}Memory \\ usage (MB)\end{tabular}} & \textbf{\begin{tabular}[c]{@{}c@{}}Latency\\ (ms)\end{tabular}} & \textbf{\begin{tabular}[c]{@{}c@{}}Speedup\\ (\%)\end{tabular}} \\ \hline

\textbf{Baseline}& \textbf{ResNet-18~\cite{he2016deep}} & 76.23 & 47.24 & 213.24 & --- \\ \hline 

\multirow{2}{*}{\textbf{\begin{tabular}[c]{@{}c@{}}Resource-friendly\\ frontend compilation\end{tabular}}} 
 & \textbf{Low-rank decomposition~\cite{yin2021towards}} & 73.73 & 15.61 & 197.53 & 7.37 \\ \cline{2-6} 

 & \textbf{Pruning~\cite{fan2023sparse}} & 71.31 & 23.91 & 146.73 & 31.19 \\ \hline

\multirow{2}{*}{\textbf{\begin{tabular}[c]{@{}c@{}}Model-adaptive\\ backend compilation\end{tabular}}} & \textbf{\begin{tabular}[c]{@{}c@{}}CoDL~\cite{jia2022codl} \\ \end{tabular}} & 76.23 & 47.24 & 189.01 & 11.36 \\ \cline{2-6} 
 & \textbf{\begin{tabular}[c]{@{}c@{}}Operator fusion~\cite{niu2021dnnfusion} \end{tabular}} & 76.23 & 47.24 & 136.66 & 35.91\\ \hline

\multirow{3}{*}{\textbf{\begin{tabular}[c]{@{}c@{}}Cross-level \\ combination\end{tabular}}} & \textbf{\begin{tabular}[c]{@{}c@{}}CoDL~\cite{jia2022codl}+Low-rank decomposition~\cite{yin2021towards} \\ \end{tabular}} & 73.72 & 15.60 & 132.96 & 37.65 \\ \cline{2-6} 
 & \textbf{\begin{tabular}[c]{@{}c@{}}CoDL~\cite{jia2022codl}+ pruning\cite{fan2023sparse} \end{tabular}} & 71.31 & 23.89 & 131.46 & 38.35\\
\cline{2-6} 
 & \textbf{CoDL\cite{jia2022codl}+pruning\cite{fan2023sparse}+operator fusion\cite{niu2021dnnfusion}+memory allocation~\cite{symons2021loma}} & 73.56 & 11.53 & 103.23 & 48.4\% \\ \hline
\end{tabular}
\vspace{-8mm}
}
\vspace{-6mm}
\end{table}

\subsubsection{Cross-level optimization opportunities}
\label{cross_phenomenon}
To evaluate the potential of cross-level optimization for enhancing parallel inference, we integrate various  techniques with CoDL~\cite{jia2022codl} on ResNet-18, as shown in Table~\ref{tab:cross}.
\textit{First}, optimizations at the frontend compilation can also improve resource utilization and accelerate DL inference. 
Low-rank decomposition reduces latency by 7.37\%, leveraging smaller-rank matrices that require less memory access. Channel pruning further reduces latency by 31.19\%, eliminating redundant channels. 
\textit{Second}, backend compilation-level optimization, such as operator fusion and CoDL parallelism, can improve resource availability, broadening the optimization space for frontend compilation while sacrificing less accuracy to meet resource budgets. 
CoDL on CPU+GPU speeds up parallel inference by 11\%, and operator fusion achieves a significant 35\% latency reduction. 
\textit{Fourth}, 
cross-level optimization optimizes the accuracy-efficiency tradeoff, with the combination of pruning, operator fusion, memory allocation, and CoDL achieving the optimal tradeoff and reducing latency by 48.4\%.

\subsection{Is parallel inference across processors always beneficial? }

\subsubsection{Performance of parallel strategies with diverse scheduling granularity}
\begin{figure}[t]
  \centering
  \includegraphics[width=0.33\textwidth]{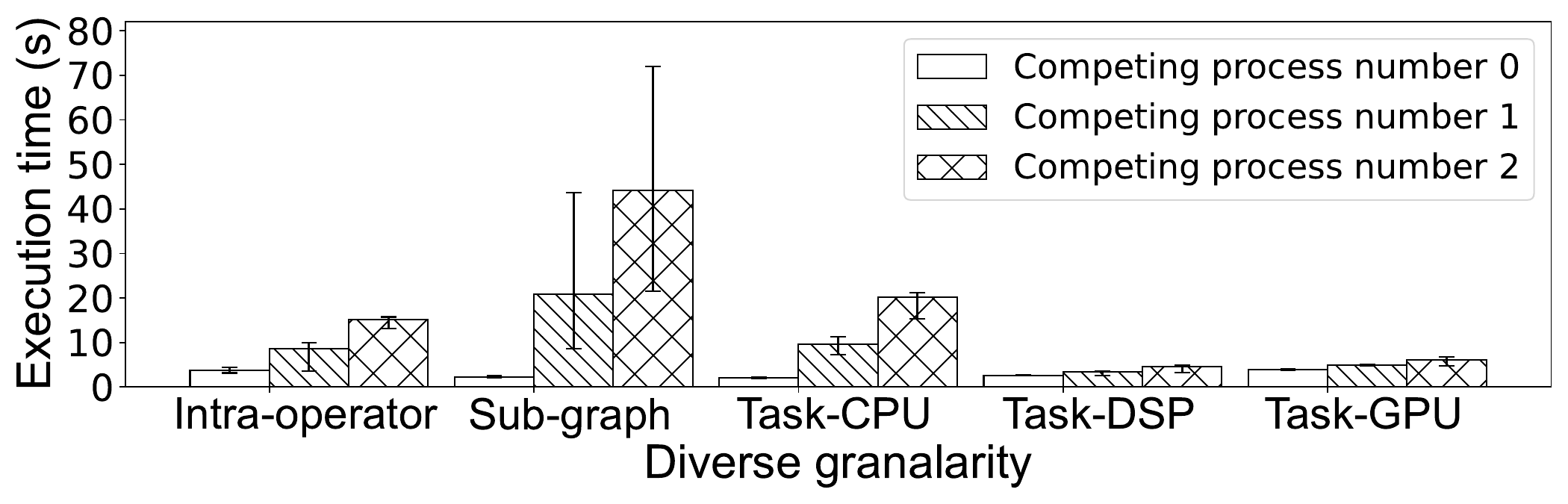}
  \vspace{-4mm}
  \caption{Impact of parallel scheduling granularity.}
  \vspace{-3mm}
  \label{fig:granularity}
\end{figure}

To test the impact of parallel scheduling granularity in real-world mobile contexts, we conducted experiments on Resnet-50 using Snapdragon 855 ($D_1$) with diverse parallel granularities, \ie \textit{\textcircled{1}} sub-graph parallelism using NN-stretch across CPU+GPU+DSP, \textit{\textcircled{2}} inter-operator parallelism using NN-stretch on CPU+GPU, \textit{\textcircled{3}} intra-operator parallelism using CoDL across CPU+GPU, and \textit{\textcircled{4}} task parallelism on CPU, GPU, and DSP using Tensorflow Lite.
Additionally, we simulate dynamic contexts by introducing varying numbers of competing processes, \ie 0, 1, and 2.
As shown in Figure \ref{fig:granularity},
\textit{first}, coarse-grained parallel granularity, \ie sub-graphs, exhibits optimal performance, \eg 19ms, when no competing processes, owing to lower data merging time. 
\textit{Second}, 
fine-grained parallel granularity, \ie intra-operator, has the lowest latency when there are 2 competing processes, attributed to the low computation load per processor. 
However, it causes process suspensions which increase data transmission delay.
With a high-competing workload, CoDL achieves the lowest latency. 
While in low workload, its latency increases by 48\% than NN-stretch. 
\textit{Third}, with a heavy workload, this impact on coarse-grained parallelism is significant.  
\textit{Fourth},
allocating computations to a single processor could, in CPU high-workload cases, outperform across heterogeneous processors due to scheduling overhead.
The scheduling time for a single GPU is 2.5 $\times$ larger than the latency reduction brought by CPU-GPU parallelism.
\textit{Fifth}, CPU are more sensitive to competing than GPU or DSP. 
With 3 competing requests, the latency on CPU increases by 10.2$\times$, while for GPU and DSP, only increases by 2.1$\times$ and 1.8$\times$.   

\textbf{Summary}. Mobile resource dynamics necessitates adaptive adjustment of parallel scheduling granularity.

\vspace{-2mm}
\begin{figure*}[t]
  \centering
  \includegraphics[width=0.7\textwidth]{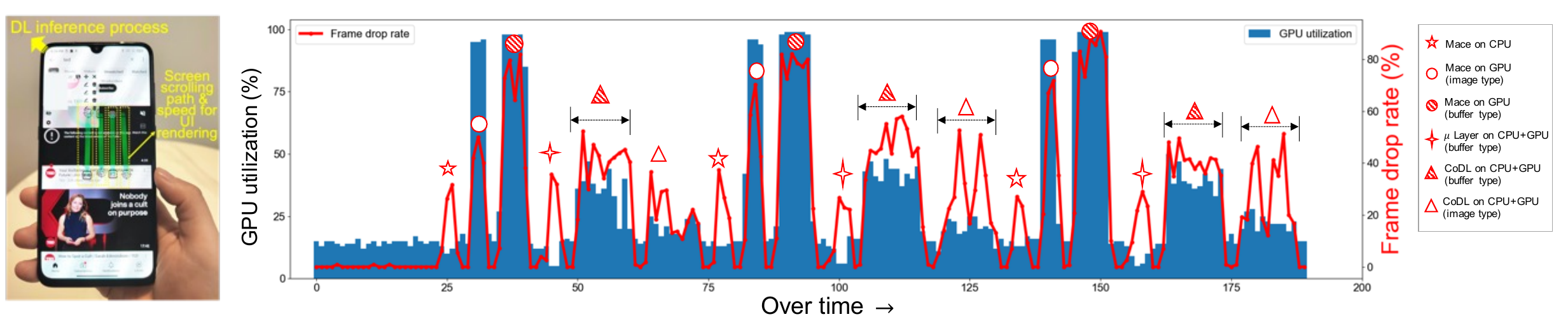}
    \vspace{-4mm}
  \caption{GPU utilization and mobile user interaction responsiveness (\eg frame drop rate) at runtime.}
  \vspace{-3mm}
  \label{fig:gpu_usage_frame_drop_rate}
  \vspace{-1mm}
\end{figure*}

\subsubsection{Performance of parallel strategies with diverse processor utilization rates} 
\label{subsubsec_processor_utilization_rates}
Maximizing resource utilization for DL inference may not always lead to optimal system performance, especially in multi-process applications.
We employ a user to simultaneously run DL inference on a smartphone while using a video playback app, \eg YouTube, showcasing the performance tradeoff between DL inference and other system operations.
Using Auto Clicker software, we establish a fixed screen scrolling path and speed, continuously scrolling the YouTube app to maintain a consistent UI rendering frame rate and workload.
In this setup, we first obtained the frame drop rate and GPU utilization when users using video playback app without any DL model inference tasks over a 2-minute period. 
We then evaluate six single-/cross-processor inference strategies over a 20-minute period with VGG-16 on device $D_1$. 
Figure \ref{fig:gpu_usage_frame_drop_rate} shows the results.
\textit{First}, excessive GPU utilization during DL inference led to significant frame drop rates in the YouTube video app refresh. 
Utilizing Mace on GPU (with buffer type) resulted in frame drops exceeding 90\%.
\textit{Second}, compared to execution on a single GPU, parallel execution results in a lower frame drop rate. 
This is because parallel execution offloads a portion of the computational burden onto the CPU, reducing the load on the GPU, which decreases the UI frame drop rate. 

\textbf{Summary}. In mobile systems, identifying optimal processor utilization levels for parallel inference is demanded to balance DL inference performance with other system tasks.

\begin{table}[]
\renewcommand{\arraystretch}{0.5}
\caption{Latency estimation errors on diverse devices.}
\vspace{-3mm}
\label{tab:dynamic-resource}
\resizebox{0.85\columnwidth}{!}{
\begin{tabular}{|c|c|ccccc|cccc|}
\hline
\multirow{4}{*}{\textbf{Model}} & \multirow{4}{*}{\textbf{Methods}}                   & \multicolumn{5}{c|}{\multirow{2}{*}{\textbf{Estimated   inference latency (ms)}}}                                                                                                                                                                                                              & \multicolumn{4}{c|}{\multirow{2}{*}{\textbf{Estimation   error (\%)}}}                                                                                                                                                       \\
                                &                                                     & \multicolumn{5}{c|}{}                                                                                                                                                                                                                                                                          & \multicolumn{4}{c|}{}                                                                                                                                                                                                        \\ \cline{3-11} 
                                &                                                     & \multicolumn{1}{c|}{\multirow{2}{*}{\textbf{$\sigma_0$ (offline)}}} & \multicolumn{1}{c|}{\multirow{2}{*}{\textbf{$\sigma_1$}}} & \multicolumn{1}{c|}{\multirow{2}{*}{\textbf{$\sigma_2$}}} & \multicolumn{1}{c|}{\multirow{2}{*}{\textbf{$\sigma_3$}}} & \multirow{2}{*}{\textbf{$\sigma_4$}} & \multicolumn{1}{c|}{\multirow{2}{*}{\textbf{$\sigma_1$↑}}} & \multicolumn{1}{c|}{\multirow{2}{*}{\textbf{$\sigma_2$↑}}} & \multicolumn{1}{c|}{\multirow{2}{*}{\textbf{$\sigma_3$↑}}} & \multirow{2}{*}{\textbf{$\sigma_4$↑}} \\
                                &                                                     & \multicolumn{1}{c|}{}                                               & \multicolumn{1}{c|}{}                                     & \multicolumn{1}{c|}{}                                     & \multicolumn{1}{c|}{}                                     &                                      & \multicolumn{1}{c|}{}                                      & \multicolumn{1}{c|}{}                                      & \multicolumn{1}{c|}{}                                      &                                       \\ \hline
\multirow{4}{*}{YOLOv2}         & \multirow{2}{*}{Mace~\cite{mace}   on GPU}          & \multicolumn{1}{c|}{\multirow{2}{*}{219}}                           & \multicolumn{1}{c|}{\multirow{2}{*}{262}}                 & \multicolumn{1}{c|}{\multirow{2}{*}{320}}                 & \multicolumn{1}{c|}{\multirow{2}{*}{270}}                 & \multirow{2}{*}{372}                 & \multicolumn{1}{c|}{\multirow{2}{*}{19}}                   & \multicolumn{1}{c|}{\multirow{2}{*}{46}}                   & \multicolumn{1}{c|}{\multirow{2}{*}{23}}                   & \multirow{2}{*}{70}                   \\
                                &                                                     & \multicolumn{1}{c|}{}                                               & \multicolumn{1}{c|}{}                                     & \multicolumn{1}{c|}{}                                     & \multicolumn{1}{c|}{}                                     &                                      & \multicolumn{1}{c|}{}                                      & \multicolumn{1}{c|}{}                                      & \multicolumn{1}{c|}{}                                      &                                       \\ \cline{2-11} 
                                & \multirow{2}{*}{CoDL~\cite{jia2022codl} on CPU+GPU} & \multicolumn{1}{c|}{\multirow{2}{*}{170}}                           & \multicolumn{1}{c|}{\multirow{2}{*}{217}}                 & \multicolumn{1}{c|}{\multirow{2}{*}{277}}                 & \multicolumn{1}{c|}{\multirow{2}{*}{263}}                 & \multirow{2}{*}{315}                 & \multicolumn{1}{c|}{\multirow{2}{*}{28}}                   & \multicolumn{1}{c|}{\multirow{2}{*}{63}}                   & \multicolumn{1}{c|}{\multirow{2}{*}{55}}                   & \multirow{2}{*}{85}                   \\
                                &                                                     & \multicolumn{1}{c|}{}                                               & \multicolumn{1}{c|}{}                                     & \multicolumn{1}{c|}{}                                     & \multicolumn{1}{c|}{}                                     &                                      & \multicolumn{1}{c|}{}                                      & \multicolumn{1}{c|}{}                                      & \multicolumn{1}{c|}{}                                      &                                       \\ \hline
\multirow{4}{*}{VGG-16}         & \multirow{2}{*}{Mace~\cite{mace}   on GPU}          & \multicolumn{1}{c|}{\multirow{2}{*}{240}}                           & \multicolumn{1}{c|}{\multirow{2}{*}{272}}                 & \multicolumn{1}{c|}{\multirow{2}{*}{312}}                 & \multicolumn{1}{c|}{\multirow{2}{*}{301}}                 & \multirow{2}{*}{464}                 & \multicolumn{1}{c|}{\multirow{2}{*}{13}}                   & \multicolumn{1}{c|}{\multirow{2}{*}{30}}                   & \multicolumn{1}{c|}{\multirow{2}{*}{25}}                   & \multirow{2}{*}{93}                   \\
                                &                                                     & \multicolumn{1}{c|}{}                                               & \multicolumn{1}{c|}{}                                     & \multicolumn{1}{c|}{}                                     & \multicolumn{1}{c|}{}                                     &                                      & \multicolumn{1}{c|}{}                                      & \multicolumn{1}{c|}{}                                      & \multicolumn{1}{c|}{}                                      &                                       \\ \cline{2-11} 
                                & \multirow{2}{*}{CoDL~\cite{jia2022codl} on CPU+GPU} & \multicolumn{1}{c|}{\multirow{2}{*}{171}}                           & \multicolumn{1}{c|}{\multirow{2}{*}{220}}                 & \multicolumn{1}{c|}{\multirow{2}{*}{307}}                 & \multicolumn{1}{c|}{\multirow{2}{*}{326}}                 & \multirow{2}{*}{387}                 & \multicolumn{1}{c|}{\multirow{2}{*}{29}}                   & \multicolumn{1}{c|}{\multirow{2}{*}{80}}                   & \multicolumn{1}{c|}{\multirow{2}{*}{91}}                   & \multirow{2}{*}{126}                  \\
                                &                                                     & \multicolumn{1}{c|}{}                                               & \multicolumn{1}{c|}{}                                     & \multicolumn{1}{c|}{}                                     & \multicolumn{1}{c|}{}                                     &                                      & \multicolumn{1}{c|}{}                                      & \multicolumn{1}{c|}{}                                      & \multicolumn{1}{c|}{}                                      &                                       \\ \hline
\multirow{4}{*}{PoseNet}        & \multirow{2}{*}{Mace~\cite{mace}   on GPU}          & \multicolumn{1}{c|}{\multirow{2}{*}{430}}                           & \multicolumn{1}{c|}{\multirow{2}{*}{442}}                 & \multicolumn{1}{c|}{\multirow{2}{*}{503}}                 & \multicolumn{1}{c|}{\multirow{2}{*}{637}}                 & \multirow{2}{*}{768}                 & \multicolumn{1}{c|}{\multirow{2}{*}{2}}                    & \multicolumn{1}{c|}{\multirow{2}{*}{17}}                   & \multicolumn{1}{c|}{\multirow{2}{*}{48}}                   & \multirow{2}{*}{79}                   \\
                                &                                                     & \multicolumn{1}{c|}{}                                               & \multicolumn{1}{c|}{}                                     & \multicolumn{1}{c|}{}                                     & \multicolumn{1}{c|}{}                                     &                                      & \multicolumn{1}{c|}{}                                      & \multicolumn{1}{c|}{}                                      & \multicolumn{1}{c|}{}                                      &                                       \\ \cline{2-11} 
                                & \multirow{2}{*}{CoDL~\cite{jia2022codl} on CPU+GPU} & \multicolumn{1}{c|}{\multirow{2}{*}{328}}                           & \multicolumn{1}{c|}{\multirow{2}{*}{335}}                 & \multicolumn{1}{c|}{\multirow{2}{*}{395}}                 & \multicolumn{1}{c|}{\multirow{2}{*}{648}}                 & \multirow{2}{*}{725}                 & \multicolumn{1}{c|}{\multirow{2}{*}{2}}                    & \multicolumn{1}{c|}{\multirow{2}{*}{20}}                   & \multicolumn{1}{c|}{\multirow{2}{*}{98}}                   & \multirow{2}{*}{121}                  \\
                                &                                                     & \multicolumn{1}{c|}{}                                               & \multicolumn{1}{c|}{}                                     & \multicolumn{1}{c|}{}                                     & \multicolumn{1}{c|}{}                                     &                                      & \multicolumn{1}{c|}{}                                      & \multicolumn{1}{c|}{}                                      & \multicolumn{1}{c|}{}                                      &                                       \\ \hline
\end{tabular}
\vspace{-3mm}
}
\vspace{-4mm}
\end{table}

\subsubsection{Offline \textit{vs.} runtime latency profiler}
%
Existing latency profilers are mainly \textit{offline}, \eg linear regression based on the number of computations (\ie FLOPs) \cite{liu2018darts}, complex black-box machine learning \cite{zhang2021nn}, and platform-aware methods (\ie concurrency) \cite{jia2022codl}.
However, the non-stationary dynamics in mobile environments widen the gap between actual and estimated latency.
To observe the impact of resource dynamics on latency, we simulate five resource conditions $\sigma_0 \sim \sigma_4$ commonly seen in mobile systems\cite{she2023accurate}: 
$\sigma_0$ (offline): run DL inference solely on GPU with 30 \textdegree C,  which affects dynamic voltage and frequency scaling (DVFS); 
$\sigma_1$: increase the mobile GPU temperature to 60 \textdegree C; $\sigma_2$: increase the mobile GPU temperature to 70 \textdegree C;
$\sigma_3$: let competing processes occupy the cache to tune the cache hit rate to be $20\%$;
$\sigma_4$: suspend half of the operators not supported by GPU to CPU for execution.
As shown in Table \ref{tab:dynamic-resource}, \textit{first}, the speedup performance under dynamic resource condition $\sigma_1 \sim \sigma_4$ significantly vary from $\sigma_0$ which has abundant resources (\ie offline settings). 
\textit{Second}, both GPU inference and CPU-GPU parallel schemes experience increased latency as the temperature rises in $\sigma_2$, due to the need for DVFS to prevent overheating.
\textit{Third}, parallel inference benefits from a higher cache hit rate, leading to lower latency than single-processor. 
Parallel inference distributes computation across processors, mitigating the impact of GPU computation slowdown caused by high-temperature throttling. 
In $\sigma_3$, increased cache competing results in frequent cache misses and memory access, increasing latency.
\textit{Fourth}, suspending operators from GPU to slower CPUs increases computation latency, and CPU-GPU data transfer incurs higher transmission latency. 
In $\sigma_4$, offloading GPU-unsupported operators to CPU results in a latency increase by 93\% for Mace and 126\% for CoDL.

\textbf{Summary}. Runtime latency profiling is desired to integrate non-stationary mobile resource dynamics.

\begin{table}[t]
\caption{Impact of data reuse in parallel inference.}
\vspace{-3mm}
\label{tab:data-reuse}
\resizebox{\columnwidth}{!}{%
\begin{tabular}{|l|c|c|c|c|c|c|}
\hline
 & \textbf{\begin{tabular}[c]{@{}c@{}}Frame-level\\ reuse\end{tabular}} & \textbf{\begin{tabular}[c]{@{}c@{}}Layer-level\\ reuse\end{tabular}} & \textbf{\begin{tabular}[c]{@{}c@{}}Latency\\ per frame (ms)\end{tabular}} & \textbf{Improvement} & \textbf{\begin{tabular}[c]{@{}c@{}}Visual quality\\ (VMAF)\end{tabular}} & \textbf{Degradation} \\ \hline
\multirow{3}{*}{\textbf{Setups}} & 0 & 0 & 99.8 & - & 49 & - \\ \cline{2-7} 
 & 1 & 1 & 93.2 & 6.6\% ↓ & 47 & 4.1\% ↓ \\ \cline{2-7} 
 & 2 & 2 & 86.2 & 13.6\% ↓ & 42 & 14.3\% ↓ \\ \hline                                                     
\end{tabular}%
\vspace{-8mm}
}
\vspace{-8mm}
\end{table}

\subsubsection{Opportunities in integrating mobile DL inference regularity into backend data reuse}
\label{subsubsec_memory_usage_lifecycle}
Mobile DL inference on continual data streams, \eg video, follows a \textit{predictable lifecycle}.
It's beneficial to determine when previous intermediate results will be accessed for the final time, and when it's safe to release resources.
We test a model on device $D_1$. 
In Table~\ref{tab:data-reuse},
the reuse in parallel inference led to a 6.6\% reduction in latency with a slight decrease, \ie 4.1\%, in visual quality (VMAF).

\section{Conclusion}

Parallel inference on heterogeneous mobile processors can boost the performance of computing-intensive DL models. 
However, mainstream mobile DL frameworks typically utilize one processor, limiting speedup. 
Despite prior efforts, the effectiveness of parallel inference in real-world mobile scenarios with diversity and dynamics is less-explored. 
We conduct a cross-level benchmark to assess potential and pitfalls of parallel DL inference on heterogeneous mobile processors. 
More insights and heuristics are needed on this topic.

\section*{Acknowledgements}
This work was supported by the National Science Fund for Distinguished Young Scholars (62025205), the National Natural Science Foundation of China (No. 62032020, 62102317), and CityU APRC grant (No. 9610633).

\bibliography{acmart}
\bibliographystyle{ACM-Reference-Format}
\end{CJK}
\end{document}